\documentclass[sigconf]{acmart}

\AtBeginDocument{%
  }

\copyrightyear{2024}
\acmYear{2024}
\setcopyright{rightsretained}
\acmConference[CIKM '24]{Proceedings of the 33rd ACM International Conference on Information and Knowledge Management}{October 21--25, 2024}{Boise, ID, USA}
\acmBooktitle{Proceedings of the 33rd ACM International Conference on Information and Knowledge Management (CIKM '24), October 21--25, 2024, Boise, ID, USA}
\acmDOI{10.1145/3627673.3679175}
\acmISBN{979-8-4007-0436-9/24/10}

\usepackage{listings}
\lstdefinestyle{ttl}{
  basicstyle=\ttfamily\footnotesize,
  breaklines=true,
  moredelim=**[is][\color{blue}]{*}{*},
}
\usepackage{balance}

\makeatletter
\gdef\@copyrightpermission{
  \begin{minipage}{0.3\columnwidth}
   \href{https://creativecommons.org/licenses/by/4.0/}{\includegraphics[width=0.90\textwidth]{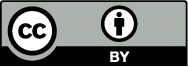}}
  \end{minipage}\hfill
  \begin{minipage}{0.7\columnwidth}
   \href{https://creativecommons.org/licenses/by/4.0/}{This work is licensed under a Creative Commons Attribution}\\{International 4.0 License.}
  \end{minipage}
  \vspace{5pt}
}
\makeatother

\begin{document}

\title{VHAKG: A Multi-modal Knowledge Graph Based on Synchronized Multi-view Videos of Daily Activities}

\author{Shusaku Egami}
\orcid{0000-0002-3821-6507}
\affiliation{%
  \institution{National Institute of Advanced Industrial Science and Technology (AIST)}
  \city{Koto-ku}
  \state{Tokyo}
  \country{Japan}
}
\email{s-egami@aist.go.jp}

\author{Takanori Ugai}
\orcid{0000-0001-5245-9719}
\affiliation{%
  \institution{Fujitsu Ltd.}
  \city{Kawasaki-shi}
  \state{Kanagawa}
  \country{Japan}}
  \email{ugai@fujitsu.com}
\affiliation{%
  \institution{National Institute of Advanced Industrial Science and Technology (AIST)}
  \city{Koto-ku}
  \state{Tokyo}
  \country{Japan}}

\author{Swe Nwe Nwe Htun}
\orcid{0000-0002-0244-2502}
\affiliation{%
  \institution{National Institute of Advanced Industrial Science and Technology (AIST)}
  \city{Koto-ku}
  \state{Tokyo}
  \country{Japan}
}
\email{swenwe.nwehtun@aist.go.jp}

\author{Ken Fukuda}
\authornote{Corresponding author}
\orcid{0000-0001-7366-1094}
\affiliation{%
  \institution{National Institute of Advanced Industrial Science and Technology (AIST)}
  \city{Koto-ku}
  \state{Tokyo}
  \country{Japan}
}
\email{ken.fukuda@aist.go.jp}

\renewcommand{\shortauthors}{Shusaku Egami, Takanori Ugai, Swe Nwe Nwe Htun, and Ken Fukuda}

\begin{abstract}
Multi-modal knowledge graphs (MMKGs), which ground various non-symbolic data (e.g., images and videos) into symbols, have attracted attention as resources enabling knowledge processing and machine learning across modalities. However, the construction of MMKGs for videos consisting of multiple events, such as daily activities, is still in the early stages. In this paper, we construct an MMKG based on synchronized multi-view simulated videos of daily activities. Besides representing the content of daily life videos as event-centric knowledge, our MMKG also includes frame-by-frame fine-grained changes, such as bounding boxes within video frames. In addition, we provide support tools for querying our MMKG. As an application example, we demonstrate that our MMKG facilitates benchmarking vision-language models by providing the necessary vision-language datasets for a tailored task.

\end{abstract}




\maketitle

\section{Introduction}
Multi-modal knowledge graphs (MMKGs)~\cite{zhu_multi-modal_2024}, which ground various non-symbolic data to symbols, have attracted attention as a resource that enables knowledge processing across modalities.
Typical MMKGs~\cite{wu_mmpedia_2023,ferrada_imgpedia_2017} are knowledge graphs (KGs) in which images are grounded to entities in the graph.
Grounding video content to entities in a KG requires solving subtasks such as event extraction, object extraction, and relation extraction, and various methods have been proposed.
In the case of a long video consisting of multiple events, such as daily activities, the sequential events need to be extracted and listed by the timeline of the events. However, these tasks are difficult to solve using current methods~\cite{zhu_multi-modal_2024}.
To the best of our knowledge, there is no available MMKG of videos consisting of such event sequences.
Furthermore, the bounding boxes of the objects within the video frames change dynamically since the information in the video changes frame by frame.
It is desirable that the MMKGs of videos can effectively represent both visual changes between each frame and contextual changes between each event that is interpreted.
Leveraging such MMKGs facilitates the construction of customized pre-training or test datasets for downstream tasks. This capability enables tailored data extraction for specific applications, such as extracting pairs of video frames and corresponding action labels before and after object state changes.

\begin{figure}[t]
\centering
\includegraphics[width=\linewidth]{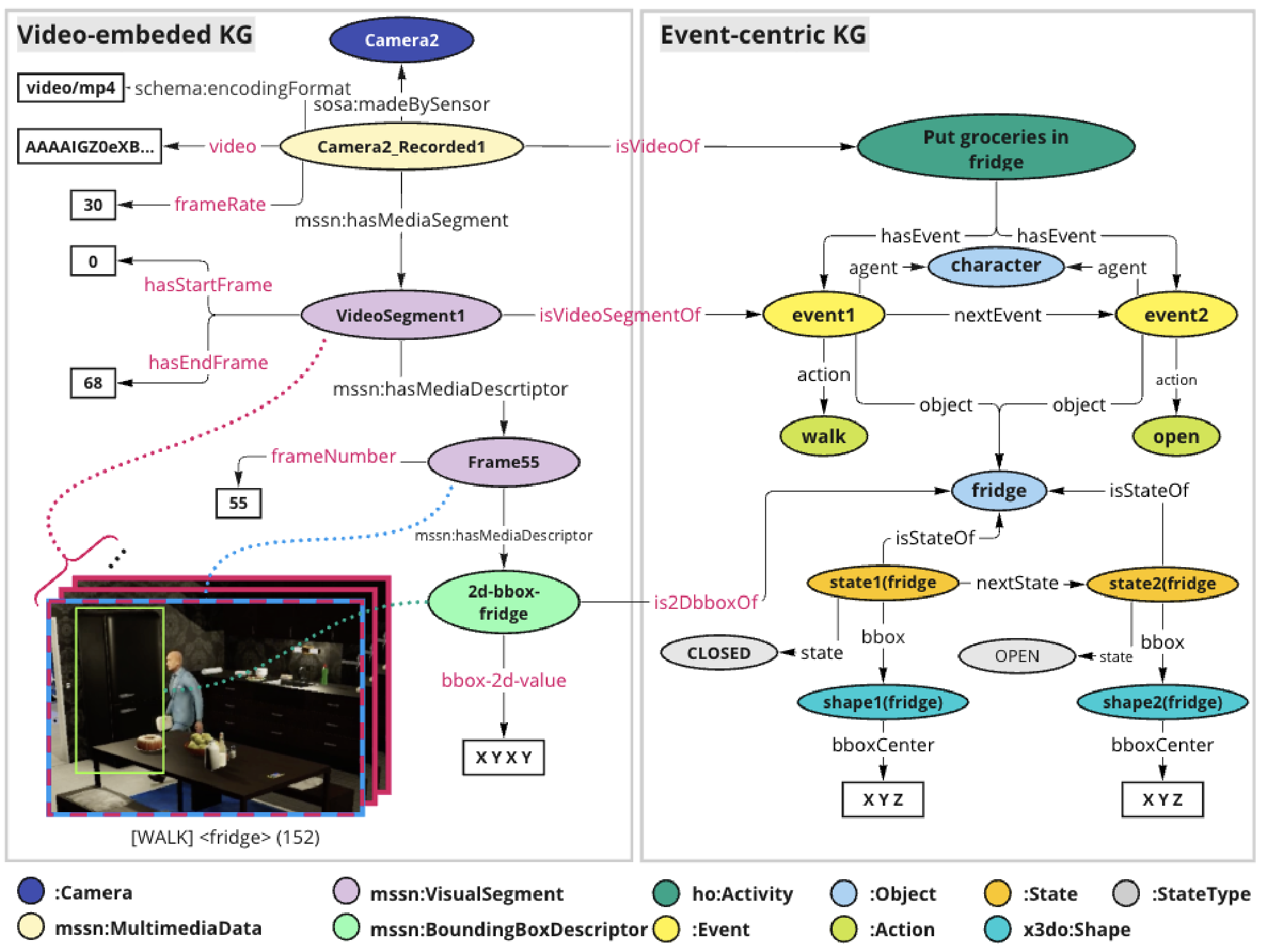}
\vspace{-6mm}
\caption{Illustration of VHAKG}
\label{fig:mmkg}
\vspace{-2mm}
\end{figure}

In this paper, we introduce a novel MMKG that integrates fine-grained events and frame-by-frame knowledge of daily activity videos.
Specifically, we first generate multi-view synchronized daily activity videos using a virtual space simulator.
Then, we construct an MMKG, called VirtualHome-AIST-KG (VHAKG), that represents event-centric and frame-by-frame knowledge, such as Figure~\ref{fig:mmkg}, based on the generated videos and our designed ontology.
Moreover, we compressed VHAKG to remove redundant triples since their data size is enormous and published it on the Web in a permanently available format.
Finally, we provide a set of tools to facilitate the use of VHAKG even by users unfamiliar with the graph query language. In addition, as a possible use case of VHAKG, we introduce the experiments that extract tailored test data sets for visual question answering (VQA) and use them to evaluate the performance of Large Vision-Language Models (LVLMs).
The main contributions of this study are summarized as follows:
\begin{itemize}
    \item \textbf{Novelty:} We constructed a novel MMKG that represents event sequences and frame-by-frame knowledge.
    \item \textbf{Availability:} We compressed our MMKG, which has a huge data size, and published it on the web in a permanently accessible format. Our MMKG is not subject to ethical review since it is artificial data.
    \item \textbf{Utility:} We provided support tools to use our MMKG and also introduced an example of creating a test set for evaluating LVLMs.
    \item \textbf{Predicted Impact:} Development of simulation-to-reality \\(sim2real) approaches~\cite{9158349,egami_synthesizing_2023,qiu_virtualhome_2023} for daily life support and the publication of MMKGs of videos, similar to our study, are expected.
\end{itemize}

\section{Related Work}
Zhu et al.~\cite{zhu_multi-modal_2024} comprehensively surveyed and summarized previous works on MMKGs.
Unfortunately, many MMKGs are not publicly available or are inaccessible.
We focus on publicly available MMKGs whose entities (i.e., nodes) link directly to image or video files.
IMGpedia~\cite{ferrada_imgpedia_2017} is an MMKG that grounds Wikimedia Commons images into DBpedia~\cite{auer_dbpedia_2007} entities.
MMpedia~\cite{wu_mmpedia_2023} is an MMKG that matches entities corresponding to images retrieved from search engines.
These MMKGs are still available because they are intended to share data using semantic web technologies.
VisionKG~\cite{yuan_visionkg_2024} is an MMKG containing bounding boxes of objects extracted from various image datasets such as MS-COCO~\cite{lin_microsoft_2014}, CIFAR~\cite{krizhevsky2009learning}, and PASCAL VOC~\cite{everingham_pascal_2010}. 
The raw dataset is not publicly available at this time, but a useful interface is available.
Our MMKG differs from these studies because it describes timelines of temporally fine-grained events in the videos and frame-by-frame bounding boxes.

Although Zhu et al.~\cite{zhu_multi-modal_2024} mentioned that the extraction of sequential events from a long video containing multiple events has not yet been addressed, the KGs of such events are constructed using different approaches from automatic event extraction.
Vizcarra et al.~\cite{vizcarra_ontology-based_2021} constructed a KG based on manual annotation data to videos. However, the KG is not publicly available because they focused on knowledge representation methods.
In our previous work~\cite{egami_synthesizing_2023}, we developed VirtualHome2KG, a framework for constructing KGs from fine-grained event data generated by VirtualHome~\cite{puig_virtualhome_2018} simulator.
The VirtualHome simulator renders human activities and outputs environmental information based on the input program data. The program data consists of multiple action steps, as shown below.
\vspace{-0.2cm}
\begin{lstlisting}[style=ttl]
Watch movie
Sit down on a couch in front of the TV. Use remote to turn on the TV.
[WALK] <couch> (275)
[SIT] <couch> (275)
[GRAB] <remote_control> (1000)
...
[WATCH] <television> (297)
\end{lstlisting}
VirtualHome2KG structures the environmental information output by VirtualHome based on an ontology to construct event-centric KGs (EKGs~\cite{guan_what_2022}), where nodes are events and entities, while edges are event-event relations, event-entity relations, and entity-entity relations.
Unlike a temporal knowledge graph (TKG)~\cite{cai_temporal_2023}, which consists of triples with timestamps, EKG is represented as an edge-labeled directed graph.
However, their entities are not directly linked to the video files because these EKGs focus on representing only the content of the videos.
In addition, the frame-by-frame knowledge, such as objects' 2D bounding boxes in each image, is not represented.
In contrast, this paper introduces a novel MMKG, which embeds synchronized video data captured by multiple cameras into entities and represents event sequences and frame-by-frame knowledge in videos.

\section{Datasets}
We describe a method for generating data on daily activities, structuring them based on an ontology, and constructing an MMKG that is practically distributable on the web through data compression.

\subsection{Data Generation}
\label{sec:data_generation}
We generated a large number of simulated videos and EKGs of a wide variety of daily activities using VirtualHome2KG.
In this step, we first added the following three new functions to the VirtualHome simulator and extended it to generate more diverse daily activity datasets frame-by-frame: (1) implementing renderable motions of various primitive actions, (2) automatically annotating 2D bounding boxes of objects in each video frame, and (3) adding synchronous camera mode with adjusted viewing angle and position.
The original VirtualHome can factually render only about 10 primitive actions, and the generated daily activity videos are not diverse enough.
Thus, we implemented 38 motions corresponding to various primitive actions, e.g., ``eat,'' ``pour,'' and ``wipe.''

In addition, we implemented a function to output the 2D bounding boxes of objects in the video frame every 5 frames.
The 2D bounding box is automatically detected by ray casting from cameras inside the environment.
Therefore, we can collect the ground truth used in computer vision tasks without any annotation work on the bounding box, which used to require much manual work.

Furthermore, we installed new fixed cameras on the diagonal of each room and increased the viewing angle of the cameras from 40 degrees to 70 degrees to capture the entire room situation.
We released the new simulator with these improvements on the web as VirtualHome-AIST\footnote{\url{https://github.com/aistairc/VirtualHome_aist}}.

We manually created over 700 various daily activity scenarios (i.e., program data described in Section 2) with reference to the existing dataset~\cite{puig_virtualhome_2018} and simulated them using VirtualHome-AIST.
Each activity was rendered simultaneously using 5 camera modes and generated over 3500 synchronized multi-view videos.
Furthermore, we integrated VirtualHome-AIST into VirtualHome2KG and constructed EKGs of over 700 daily activities.
These EKGs are integrated into video-embedded KGs described in Section~\ref{sec:mmkg_construction}.

\begin{figure}[t]
\centering
\includegraphics[width=\linewidth]{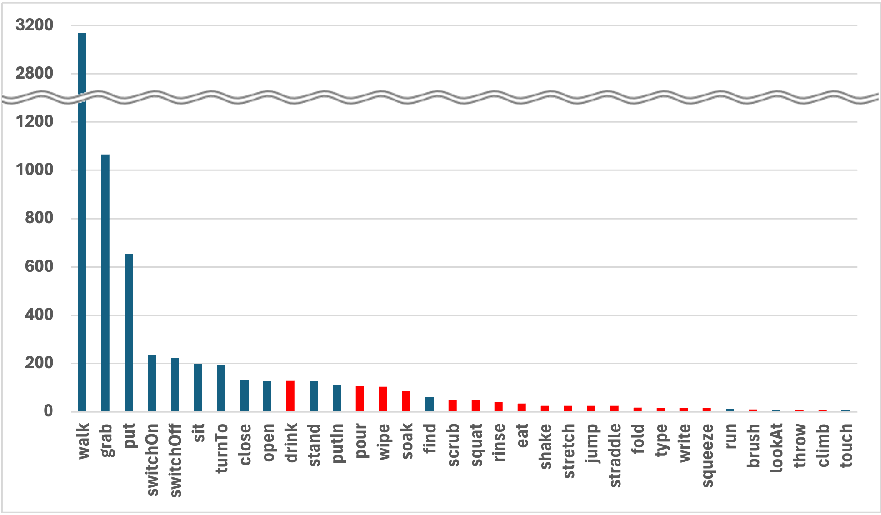}
\vspace{-6mm}
\caption{Number of actions (red means new actions that became executable by this study)}
\label{fig:actions}
\vspace{-4mm}
\end{figure}

Figure~\ref{fig:actions} shows the number of actions included in the generated video.
An activity consists of an average of 10.2 events, and each event has one action.

\subsection{MMKG Construction}
\label{sec:mmkg_construction}
Our MMKG is defined as a graph $\mathcal{G}=\{\mathcal{E, R, L, T}\}$, where $\mathcal{E, R, L}$ are set of entities, relations, and literal values, and $\mathcal{T}~\subseteq~\mathcal{E~\times~R~\times~}$
$\mathcal{(E~\bigcup~L)}$ are sets of triples. A set of literal values $\mathcal{L=\{L_{K}, L_{M}\}}$ denotes that $\mathcal{L_{K}}$ is the set of the KG's literal values and $\mathcal{L_{M}}$ is the set of multi-modal data.

We first designed the schema of VHAKG with reference to the existing ontologies.
We interpreted the videos captured by multiple cameras installed in VirtualHome-AIST as sensor data.
Thus, we reused MSSN-Onto~\cite{angsuchotmetee_mssn-onto_2020}, which is an extension of the Semantic Sensor Network Ontology~\cite{taylor_semantic_nodate} for multimedia content, and extended the VirtualHome2KG's ontology.
Figure~\ref{fig:mmkg} shows an example of the modeling of VHAKG to be constructed in this study.
The right side of the figure shows the EKG constructed in Section~\ref{sec:data_generation}.
The left side of the figure is the KG described in this section.
In the EKG, ``Activity'' corresponds to the whole video, and ``Event'' corresponds to ``VideoSegment.''
The 2D bounding box links to the corresponding ``Object'' of the EKG.
The video data is embedded as a literal value encoded in base64. 
Consequently, video files are free of broken links and reference errors, ensuring their permanent availability and sharing.

VirtualHome-AIST outputs a frame-by-frame image, a file recording the start and end frame numbers of each action step, and 2D bounding boxes every 5 frames.
We integrate these data and transform them into a KG in RDF format based on the designed schema using Python scripts with RDFLib~\cite{noauthor_rdflib_nodate}.

\subsection{MMKG Compression}
\label{sec:mmkg_compression}

\subsubsection{Video compression}
We first created a KG with all images embedded every five frames; however, we found that the data size became too large to share practically.
Thus, we used MPEG~\cite{le1991mpeg} compression to reduce the data size significantly.
Each video frame entity has a frame number instead of having a base64 value, and the video entity has a base64 value for the entire compressed video.
This allows the use of well-known tools such as FFmpeg~\cite{noauthor_ffmpeg_nodate} to recover and extract arbitrary frame images from VHAKG.

\subsubsection{Removing redundant triples}
Constructing a KG, as shown in Figure~\ref{fig:mmkg}, will create redundant triples about the 2D bounding boxes.
We reduced the number of entities and triples by referring to the previous entities if the current 2D bounding boxes have not changed since the previous frame.

\subsubsection{Results}
Table~\ref{table:kg} shows the number of triples and data size of VHAKG.
In image-embedded KG, each image was converted to JPEG with a quality of 90 using Pillow~\cite{noauthor_pillow_nodate}.
As a result, our approach significantly compressed the number of triples and data size, and made it possible to share them securely in a research data repository\footnote{For example, Zenodo's size limit is currently 50 GB.}.
VHAKG, consisting of the compressed video-embedded KG and EKG, is available at Zenodo\footnote{\url{https://doi.org/10.5281/zenodo.11438499}}.

\begin{table}[]
\caption{Number of triples and data size of constructed KG. The round brackets mean the compression ratio. ImageKG is image-embedded KG, and VideoKG is video-embedded KG.}
\vspace{-3mm}
\label{table:kg}
\begin{tabular}{lll}
\hline
    & \# of triples & Size [GB] \\ \hline
ImageKG &  134,945,485 ( - ) & 62.0 ( - ) \\ 
VideoKG  & 131,786,665 (97.7\%) & 17.3 (27.9\%) \\ 
VideoKG (compressed) & 37,646,681 (27.9\%) & 12.5 (20.0\%)  \\ \hline
\end{tabular}
\vspace{-2mm}
\end{table}

\section{Applications}
\label{sec:application}

\subsection{Support tools}

We have developed and released support tools (GUI and command-line)\footnote{\url{https://github.com/aistairc/vhakg-tools}} to make VHAKG accessible to users who are unfamiliar with SPARQL~\cite{world2013sparql}.
The GUI displays the specified videos and images and is provided as Docker Compose~\cite{noauthor_docker_0100}. 
The back-end system automatically loads VHAKG by launching GraphDB~\cite{noauthor_ontotext_nodate} as a triple store on the local machine.
The GUI executes template-based SPARQL queries that search for videos matching the conditions specified in the UI and restore and display the videos from their base64 values. 
The seek bar moves to the position of the specified frame.
In addition, the command-line tool can extract videos and images and save them with annotation labels on a local machine.

\subsection{Example Benchmarking of LVLMs}

SPARQL querying enables users to extract videos, video frames, entities, and their labels from VHAKG.
Thereby, users can create customized image and video annotation datasets for specific use cases.
As a demonstration of VHAKG, we designed a new VQA task, created a test dataset, and conducted an example experiment.

\subsubsection{Task design}

\begin{figure}[t]
\centering
\includegraphics[width=\linewidth]{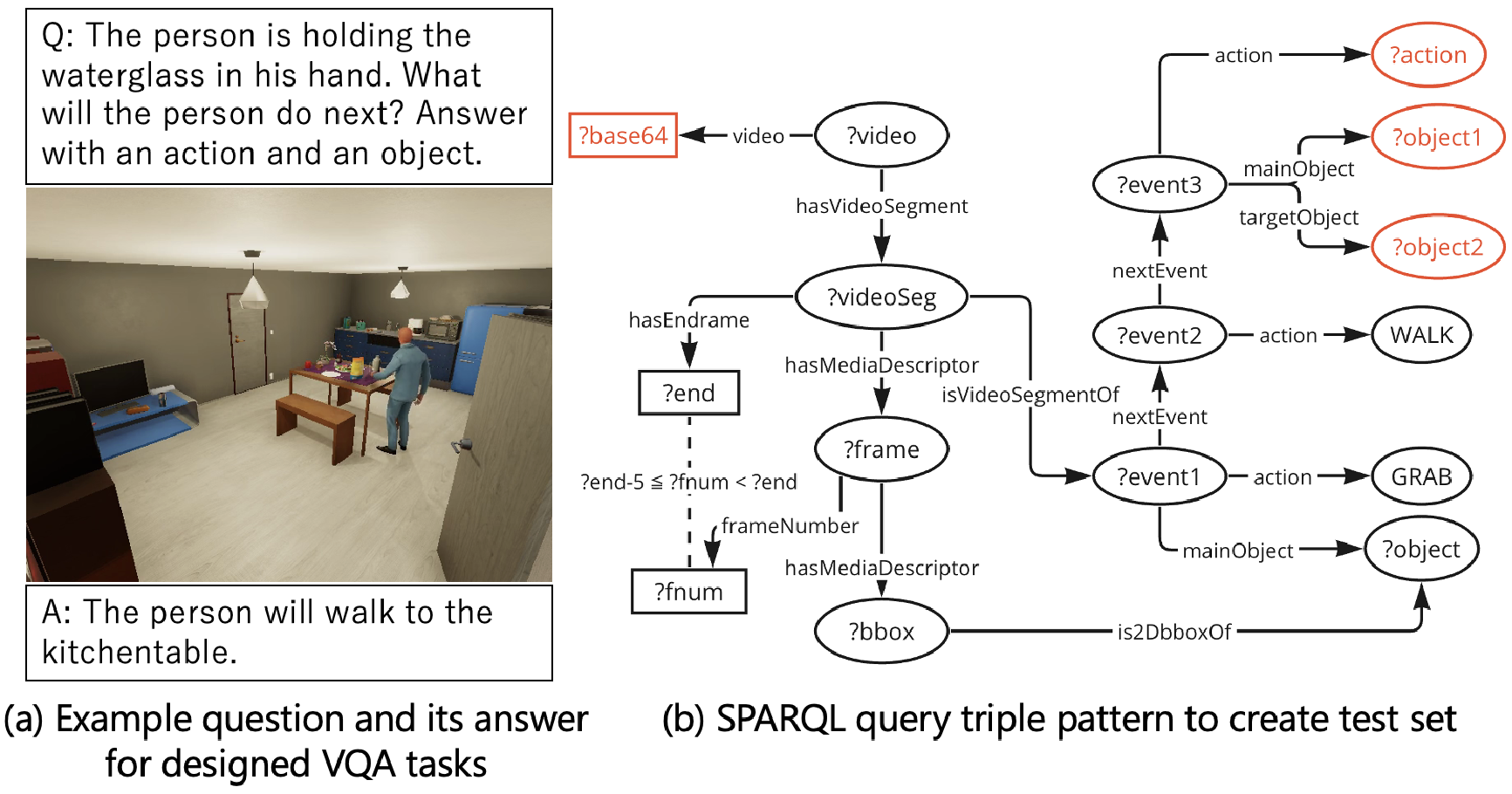}
\vspace{-6mm}
\caption{Example question and query pattern}
\label{fig:qa}
\vspace{-5mm}
\end{figure}

We designed a task to understand a character's daily activities from an input image and predict and explain the character's next action.
An example of input data is a single image and a question, as shown in Figure~\ref{fig:qa}(a).
This image is extracted from the daily activity video included in VHAKG.
The models are required to understand the meaning of this input image and explain what action the character will take next.
Therefore, this task is more difficult than existing vision-language tasks, such as caption generation.

\subsubsection{Dataset preparation}

The evaluation dataset should be able to predict the next action from the image features to some extent.
Thus, we extracted the image immediately after the character grabbed something as the question data and extracted the next action and the target objects as the correct answer labels.
Figure~\ref{fig:qa}(b) shows the triple pattern of SPARQL queries to obtain these test sets.
Red entities and literals are extracted.
This triple pattern queries the event's action following the event that the character grabs something and walks to somewhere, the object, the video, and the frame number.
We then created short sentences based on the extracted actions and objects, as shown in Figure~\ref{fig:qa}(a), based on a simple template.
We extracted 100 pairs of data as the test set and another 5 pairs of data as samples for few-shot learning.
Such data extraction is possible because VHAKG comprehensively represents video data, event-centric knowledge, and frame-by-frame knowledge.

\subsubsection{Experiments}
\begin{figure}[t]
\centering
\includegraphics[width=\linewidth]{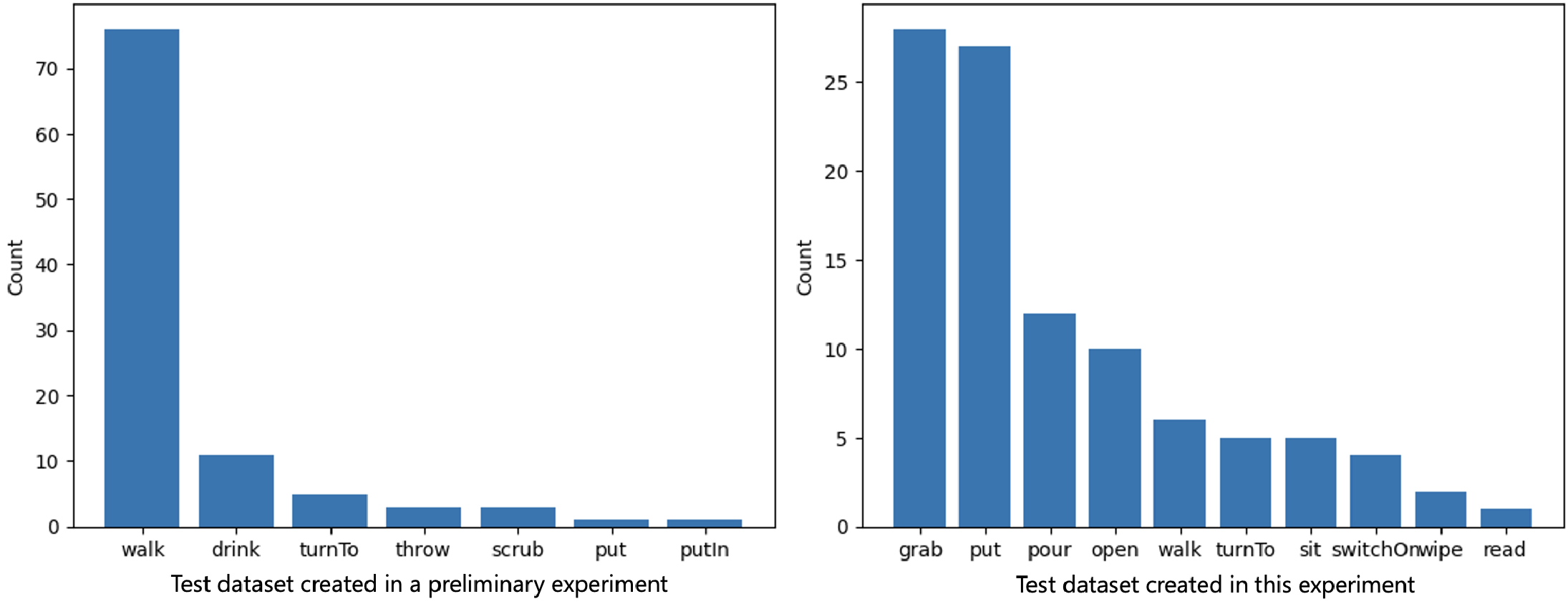}
\vspace{-7mm}
\caption{Distribution of actions in test datasets}
\label{fig:testset}
\vspace{-5mm}
\end{figure}

In our preliminary experiment, we extracted events after the ``grab'' event and found that 76 out of 100 events were ``walk,'' which was highly biased. 
In contrast, we reduced the bias in this experiment by extracting events after the ``walk'' following ``grab,'' as shown in Figure~\ref{fig:testset}.
In this way, it is possible to create a test set with reduced bias by querying VHAKG.

Note that the purpose of this section is to show that VHAKG can produce benchmark datasets for VQA tasks and that this study does not aim to develop or compare LVLMs.
We show an example of an experiment to evaluate whether LVLMs can answer questions shown in Figure~\ref{fig:qa}(a) by In-context Learning.
We selected GPT-4V (gpt-4-1106-vision-preview~\cite{noauthor_new_nodate}) and GPT-4o (gpt-4o-2024-05-13~\cite{noauthor_hello_nodate}) as LVLMs.
The evaluation was performed in zero-shot learning and 5-shot learning.
All output token sizes were set to 50.
We employed BLEU~\cite{papineni_bleu_2002}, ROUGE-1~\cite{lin_rouge_2004}, and METEOR~\cite{banerjee_meteor_2005} as evaluation metrics to calculate the similarity between the short sentences output and the correct answers.
Table~\ref{table:results} shows the evaluation results.

\begin{table}[t]
\caption{Evaluation results of VQA task}
\label{table:results}
\vspace{-3mm}
\begin{tabular}{lllll}
\hline
       Model & Method & BLEU & ROUGE-1 & METEOR \\ \hline
GPT-4V & Zero-shot & 0.249  & 0.354    & 0.271   \\
 & 5-shot  & 0.470 & 0.544   & 0.484  \\
GPT-4o & Zero-shot & 0.0808 & 0.232 & 0.111 \\ 
 & 5-shot  & 0.481 & 0.565   & 0.497  \\ 
\hline
\end{tabular}
\vspace{-5mm}
\end{table}

As a result, GPT-4 and GPT-4o could hardly predict the correct answer in the zero-shot learning of this task.
Therefore, these results imply that the new benchmark dataset created from VHAKG was an unknown set of images and labels that had not yet been trained on GPT-4V and GPT-4o.
In contrast, the accuracy improved in few-shot learning.
Therefore, these results imply that the extracted sample data are good examples for this task, and GPT-4V and GPT-4o can learn this context to some extent.

\section{Conclusion}

In this study, we provided VHAKG, which is a novel MMKG based on multi-view videos of daily activities consisting of multiple events.
Moreover, we presented VirtualHome-AIST, support tools of VHAKG, and an example experiment.
VHAKG integrated different modalities by embedding videos as literal values within the KG. 
By compressing videos and removing redundant triples, we reduced the data size and made it permanently available through a reliable research repository.
Moreover, VHAKG can contribute to the creation of benchmark datasets and pre-training datasets for LVLMs since VHAKG represents both event sequences and frame-by-frame visual changes. 
The remaining issues in this study are increasing the variety of videos and linking to other KGs. In the future, we plan to generate multi-agent and egocentric videos and link VHAKG to real visual dataset KGs~\cite{yuan_visionkg_2024,yamamoto2023towards} for sim2real tasks.

\begin{acks}
This paper is based on results obtained from a project, JPNP20006, commissioned by the New Energy and Industrial Technology Development Organization (NEDO), and JSPS KAKENHI Grant Numbers JP22K18008 and JP23H03688.
\end{acks}

\bibliographystyle{ACM-Reference-Format}
\balance
\bibliography{ref}




\end{document}